\documentclass{article}

\PassOptionsToPackage{numbers}{natbib}

\usepackage[final]{nips_2018}

\usepackage[utf8]{inputenc} 
\usepackage[T1]{fontenc}    
\usepackage{hyperref}       
\usepackage{url}            
\usepackage{booktabs}       
\usepackage{amsfonts}       
\usepackage{nicefrac}       
\usepackage{microtype}      
\usepackage{amssymb}
\usepackage{graphicx}
\usepackage{subfig}
\usepackage{color}
\usepackage{footnote}
\makesavenoteenv{tabular}
\makesavenoteenv{table}

\title{Cross-Modulation Networks For Few-Shot Learning}

\author{
  Hugo Prol\textsuperscript{1}, Vincent Dumoulin\textsuperscript{2}, and
  Luis Herranz\textsuperscript{1} \\
  \textsuperscript{1}Computer Vision Center, Universitat Autònoma de Barcelona \\
  \textsuperscript{2}Google Brain \\
  \texttt{hugo.prol.pereira@gmail.com, vdumoulin@google.com, lherranz@cvc.uab.es}
}

\begin{document}

\maketitle

\begin{abstract}
  A family of recent successful approaches to few-shot learning relies
  on learning an embedding space in which predictions are made by
  computing similarities between examples. This corresponds to combining
  information between support and query examples at a very late
  stage of the prediction pipeline. Inspired by this observation, we
  hypothesize that there may be benefits to combining the information
  at various levels of abstraction along the pipeline. We present an
  architecture called \emph{Cross-Modulation Networks} which allows
  support and query examples to interact throughout the feature
  extraction process via a feature-wise modulation mechanism. We adapt the
  Matching Networks architecture to take advantage of these interactions
  and show encouraging initial results on miniImageNet in
  the 5-way, 1-shot setting, where we close the gap with state-of-the-art.
\end{abstract}

\section{Introduction}

Recent deep learning successes in areas such as image
recognition~\cite{simonyan14vgg,he2016resnet,hu17squeeze}, machine
translation~\cite{wu16googlemt,lample18phrase}, and speech
synthesis~\cite{oord2016wavenet} rely on large amounts of data and
extensive training. In contrast, humans excel in learning new concepts
with very little supervision. Few-shot learning aims to close this gap
by training models that can generalize well from few labeled examples.

Several approaches have been proposed to tackle the few-shot learning
problem, such as learning an initialization suitable to a small number
of parameter updates when learning on a new problem with a small
amount of
data~\cite{finn2017maml,nichol18reptile,kim2018bayesian,finn18platipus,grant18llama,rusu18leo},
learning an embedding space in which examples are compared to make
predictions~\cite{vinyals2016matching,snell2017proto,sung17relnet2,oreshkin18tadam},
learning the optimization algorithm which produces the final
model~\cite{ravi2017optim}, or learning a model which adapts to new
problems through external memories or attention
mechanisms~\cite{santoro2016meta,munkhdalai17condshift,shyam2017attentive,mishra2018simple}.
In particular, the approach which consists in learning an embedding
space represents a simple yet effective solution to few-shot learning.
This approach incorporates the inductive bias that examples should be
compared at an abstract level of representation, which may be overly
restrictive in cases where intermediate representations encode
information useful to classification.

In this work we explore an extension applicable to metric learning
architectures which allows the interaction of support and query
examples at each level of abstraction. These interactions are
implemented through a \emph{cross-modulation mechanism} and enable the
network to modify the intermediate features of the compared examples
to produce better final representations and consequently a more robust
metric space. Preliminary results on miniImageNet extending Matching
Networks with this approach yield a performance comparable with
state-of-the-art architectures of similar sizes.

\section{Method}
\label{sec:method}

\paragraph{Few-shot learning and the episodic framework} We adopt the
{\em episodic approach} proposed by \cite{vinyals2016matching} and
the nomenclature introduced in \cite{ren2018meta}. We partition classes
into sets of $C_{train}$ and $C_{test}$ classes and train the model on
$K$-shot, $N$-way episodes where the model learns from a {\em support set}
$\mathcal{S} = \{(\mathbf{x}_i, y_i)\}_{i=1}^{NK}$ containing $K$ examples
for each of the $N$ classes  (which are sampled from $\mathcal{C}_{train}$)
and is required to generalize to a held-out {\em query set}
$\mathcal{Q} = \{(\mathbf{x}_j^*, y_j^*)\}_{j=1}^{T}$. The model is evaluated on
episodes constructed from $\mathcal{C}_{test}$.

\paragraph{Matching Networks}

In their simplest formulation, Matching Networks express
$p(y^* \mid \mathbf{x}^*, \mathcal{S})$ by having each example
in $\mathcal{S}$ cast a weighted vote, i.e.
\begin{equation}
  \label{eq:softmax}
  p(y^*=c \mid \mathbf{x}^*, \mathcal{S}) =
      \sum_{i=1}^{NK} h(\mathbf{x}^*, \mathbf{x}_i) 1_{y_i = c}, \quad
  h(\mathbf{x}^*, \mathbf{x}_j) =
      \frac{\exp(\mathrm{cosine}(f(\mathbf{x}^*), f(\mathbf{x}_i)))}
           {\sum_{j = 1}^{NK} \exp(\mathrm{cosine}(f(\mathbf{x}^*), f(\mathbf{x}_j)))},
\end{equation}
where $f(\cdot)$ is a parametrized feature extractor, and $h(\cdot, \cdot)$
computes the softmax over cosine similarities to the query example. Our
implementation of $f(\cdot)$ follows the standard architecture used in
the literature: a convolutional neural network with four blocks, each formed
of a $3 \times 3$ convolution with 64 filters followed by batch
normalization~\cite{ioffe15batchnorm}, a ReLU activation function, and
a $2 \times 2$ max pooling operation.

\begin{figure}[t]
  \centering
  \subfloat[FiLM generator\label{fig:filmgen}]{\includegraphics[width=0.2\textwidth]{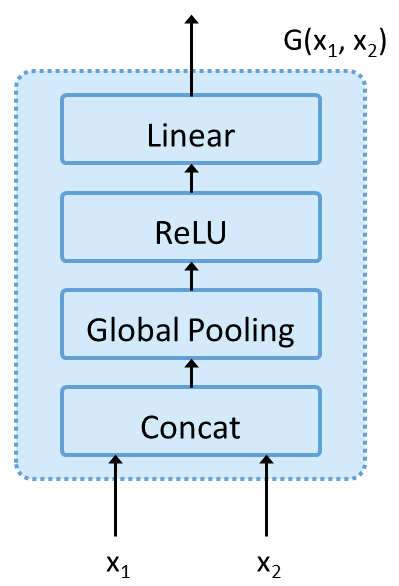}}\qquad
  \subfloat[Cross-modulation mechanism\label{fig:xmodblock}]{\includegraphics[width=0.32\textwidth]{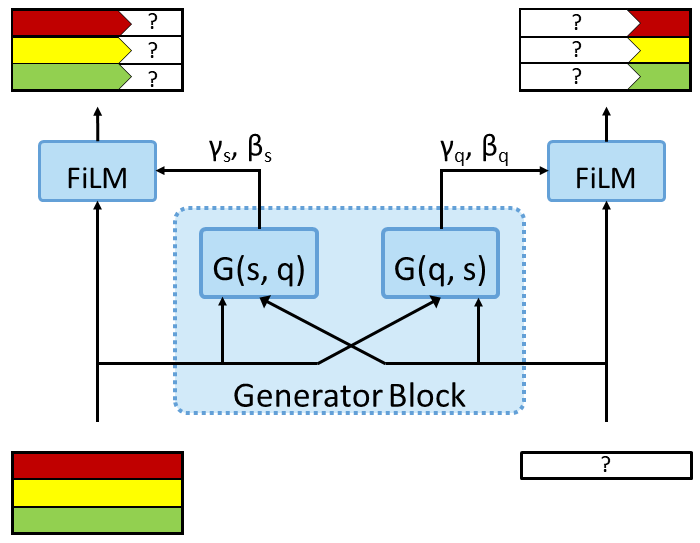}}\qquad
  \subfloat[Final convolutional block\label{fig:convblock}]{\includegraphics[width=0.32\textwidth]{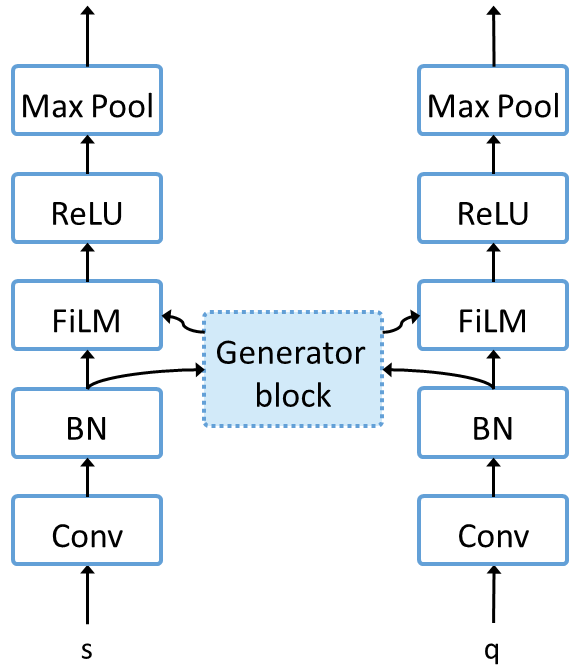}}\qquad
  \caption{Architecture of the cross-modulation mechanism.}
  \label{fig:pairwise_mod}
\end{figure}

\paragraph{Cross-modulation through FiLM layers}
Feature-wise Linear Modulation (FiLM) \cite{perez2017film} allows to
\emph{modulate} the inner representation of a network by
applying a feature-wise affine transformation. Let
$x~\in~\mathbb{R}^{H \times W \times C}$ be the output of a
convolutional layer for a given example, then a FiLM layer has the
form $\mathrm{FiLM}(x) = \gamma_z \odot x + \beta_z$, where $\odot$ denotes
the Hadamard product and $\gamma_z,\beta_z \in \mathbb{R}^C$ are the
\emph{FiLM parameters}, computed by the \emph{FiLM generator} $G(z)$
which takes the conditioning input $z$.

We introduce interactions between
query and support examples using FiLM layers (see
Figure~\ref{fig:xmodblock}). The FiLM generator implements
$G(x_1, x_2) = \varphi\left([x_1, x_2]\right) W + b$, where the weight matrix
$W \in \mathbb{R}^{2C \times 2C}$ and bias
$b \in \mathbb{R}^{2C}$ are learnable parameters,
$[\cdot, \cdot]$ denotes channel-wise concatenation, and
$\varphi: \mathbb{R}^{H \times W \times 2C} \to \mathbb{R}^{2C}$ represents
global average pooling followed by ReLU (Figure~\ref{fig:filmgen}).

Our specific FiLM layer implementation is formulated as
$\mathrm{FiLM}(x) = (1 + \gamma_0\gamma_z) \odot x + \beta_0\beta_z$,
as suggested by Oreshkin~et~al.~\cite{oreshkin18tadam}. We apply an $L_1$
regularization penalty to $\gamma_0$ and $\beta_0$ to enforce sparsity in
the modulation, under the assumption that some features are more relevant
than others in the modulation process. The input of the FiLM generator
$G$ is a batch of support-query example pairs resulting from the Cartesian
product of the support and query batches. Note that in general
$G(s_i, q_i) \neq G(q_i, s_i)$.

Previous work such as \cite{oreshkin18tadam,munkhdalai17condshift} also makes
use of feature-wise modulation in a few-shot learning context; our approach
differentiates itself by considering pairwise interactions where both the support
and the query example are used to predict the modulating parameters, and
prediction happens locally at multiple levels of abstraction rather than via
high-level features extracted from the support set.

\paragraph{Cross-Modulation Networks}

We extend Matching Networks by augmenting the feature extractor with
our proposed cross-modulation mechanism. We use two embedding
functions with shared parameters to encode separate batches of support
and query examples. Their first convolutional block is identical to
that of the baseline, while the other three blocks are
cross-modulated, as illustrated in Figure~\ref{fig:convblock}. We
observed empirically that cross-modulation of the first block did not
improve the results, and we decided not to cross-modulate it to save
on computational overhead.

\section{Experiments}
\label{sec:experiments}

\paragraph{Setup} We experiment on the miniImageNet
dataset~\cite{vinyals2016matching} using the splits proposed
by Ravi~and~Larochelle~\cite{ravi2017optim}. We train using the Adam
optimizer~\cite{kingma14adam}; the initial learning rate is set to $0.001$
and halved every $10^5$ episodes. For the {\em 5-way, 1-shot}
setup we use 15 query examples per episode when training Matching
Networks but use 5 query examples per episode when training Cross-Modulation
Networks, due to computational constraints. We set the L1 factor for
the $\gamma_0$ and $\beta_0$ post-multipliers to $0.001$ following
cross-validation. Test accuracies are averaged over 1000 episodes using
15 query examples per episode, and we report 95\% confidence intervals.

\begin{table}[t]
\setlength{\tabcolsep}{3pt}
  \begin{minipage}{.62\linewidth}
  \centering
  \caption{\label{tab:miniimagenet} Test accuracy on miniImageNet (\%).}
  \begin{tabular}{lcc}
    \toprule
    \textbf{Model} & \textbf{5-way 1-shot} & \textbf{5-way 5-shot} \\
    \midrule
    Meta-Learner LSTM \cite{ravi2017optim} & 43.44 $\pm$ 0.77 & 60.60 $\pm$ 0.71 \\
    MAML \cite{finn2017maml} & 48.70 $\pm$ 1.84 & 63.11 $\pm$ 0.92 \\
    Matching Networks\footnote{Our re-implementation.} \cite{vinyals2016matching} & 49.39 $\pm$ 0.62 & 66.16 $\pm$ 0.68 \\
    Prototypical Networks \cite{snell2017proto} & 49.42 $\pm$ 0.78 & \textbf{68.20 $\pm$ 0.66} \\
    REPTILE \cite{nichol18reptile} & 49.97 $\pm$ 0.32 & 65.99 $\pm$ 0.58 \\
    PLATIPUS \cite{finn18platipus} & 50.13 $\pm$ 1.86 & - \\
    Relation Nets \cite{sung17relnet2} & \textbf{50.44 $\pm$ 0.82} & 65.32 $\pm$ 0.70 \\
    Meta-SGD \cite{li17metasgd} & \textbf{50.47 $\pm$ 1.87} & 64.03 $\pm$ 0.94 \\
    \midrule
    Cross-Modulation Nets & \textbf{50.94 $\pm$ 0.61} & 66.65 $\pm$ 0.67\\
    \bottomrule
  \end{tabular}
  \end{minipage}%
\begin{minipage}{.38\linewidth}
  \setlength{\tabcolsep}{6pt}
  \centering
  \caption{\label{tab:noise} Adding per-block noise in cross-modulation
  (5-way 1-shot on miniImageNet, ticks denoting noise introduction).}
  \begin{tabular}{cccc}
    \toprule
    \textbf{2} & \textbf{3} & \textbf{4} & \textbf{Accuracy (\%)} \\
    \midrule
    {} & {} & {} &  50.94 $\pm$ 0.61 \\
    \checkmark & {} & {} &  47.68 $\pm$ 0.58 \\
    {} & \checkmark & {} &  49.28 $\pm$ 0.61 \\
    {} & {} & \checkmark &  46.50 $\pm$ 0.58 \\
    \checkmark & \checkmark & \checkmark &  42.97 $\pm$ 0.56 \\
    \bottomrule
  \end{tabular}

   \end{minipage}%
\end{table}

\paragraph{Cross-modulation helps increase test accuracy}

Table~\ref{tab:miniimagenet} compares our proposed approach with
results published for comparable network architectures.  We first note
that our Matching Networks implementation exhibits a stronger accuracy
($49.39 \pm 0.62\%$) than what is usually reported, which is due to
our use of the ``unnormalized'' cosine similarity used in the original
implementation,\footnote{We confirmed this through personal
  communication with the paper's authors.} i.e.
\begin{equation}
    \mathrm{cosine_u}(\mathbf{x}^*, \mathbf{x}_i) = \mathbf{x}^* \cdot \mathbf{x}_i \|\mathbf{x}^*\|^{-1}.
\end{equation}
We hypothesize this is directly linked to the metric scaling properties
discussed in \cite{oreshkin18tadam}.

Applying our proposed cross-modulation architecture to the baseline
Matching Networks model trained in the {\em 5-way, 1-shot} setting
increases its accuracy to $50.94 \pm 0.61\%$ --- a $1.55\%$ increase over
the corresponding baseline --- and places it on-par with other SOTA
approaches. In the {\em 5-way, 5-shot} setting, our implementation of
Matching Networks is also very competitive, and augmenting it with our proposed
cross-modulation mechanism appears to improve its accuracy, albeit more modestly ($0.59\%$). It could be that cross-modulation is best suited to the very
low-data regime and offers diminishing returns as the number of examples per
class increases, but a more thorough investigation is required to draw definitive
conclusions. Future work can also address the design of more efficient and
effective ways to exploit interactions in the few-shot case.

\paragraph{The model takes advantage of cross-modulation}

\autoref{tab:miniimagenet} shows an improvement over the Matching
Networks baseline, which we attribute to the interaction between
support and query examples via our proposed cross-modulation
mechanism. To verify that this interaction is necessary for the
performance of the trained network, we compare the test accuracy of
the model in normal conditions to its performance when the modulation
mechanism is distorted with random noise. This can be achieved
multiplying the $\gamma_0$ and $\beta_0$ respectively by new terms
$\gamma_{noise}$ and $\beta_{noise}$, drawn from a normal
distribution. Comparing the first and last rows in \autoref{tab:noise}
we observe a $7.97\%$ drop in test accuracy when distorting all the
cross-modulation blocks, which indicates that the model has learned to
take advantage of them during training. We employed
$\mathcal{N}(1, 0.3)$ for all the experiments in \autoref{tab:noise},
but similar conclusions are reached from other standard deviation values.

\begin{figure}[t]
  \centering
  \subfloat[$\gamma_0$ and $\beta_0$ distributions
  ]{\includegraphics[width=0.24\textwidth]{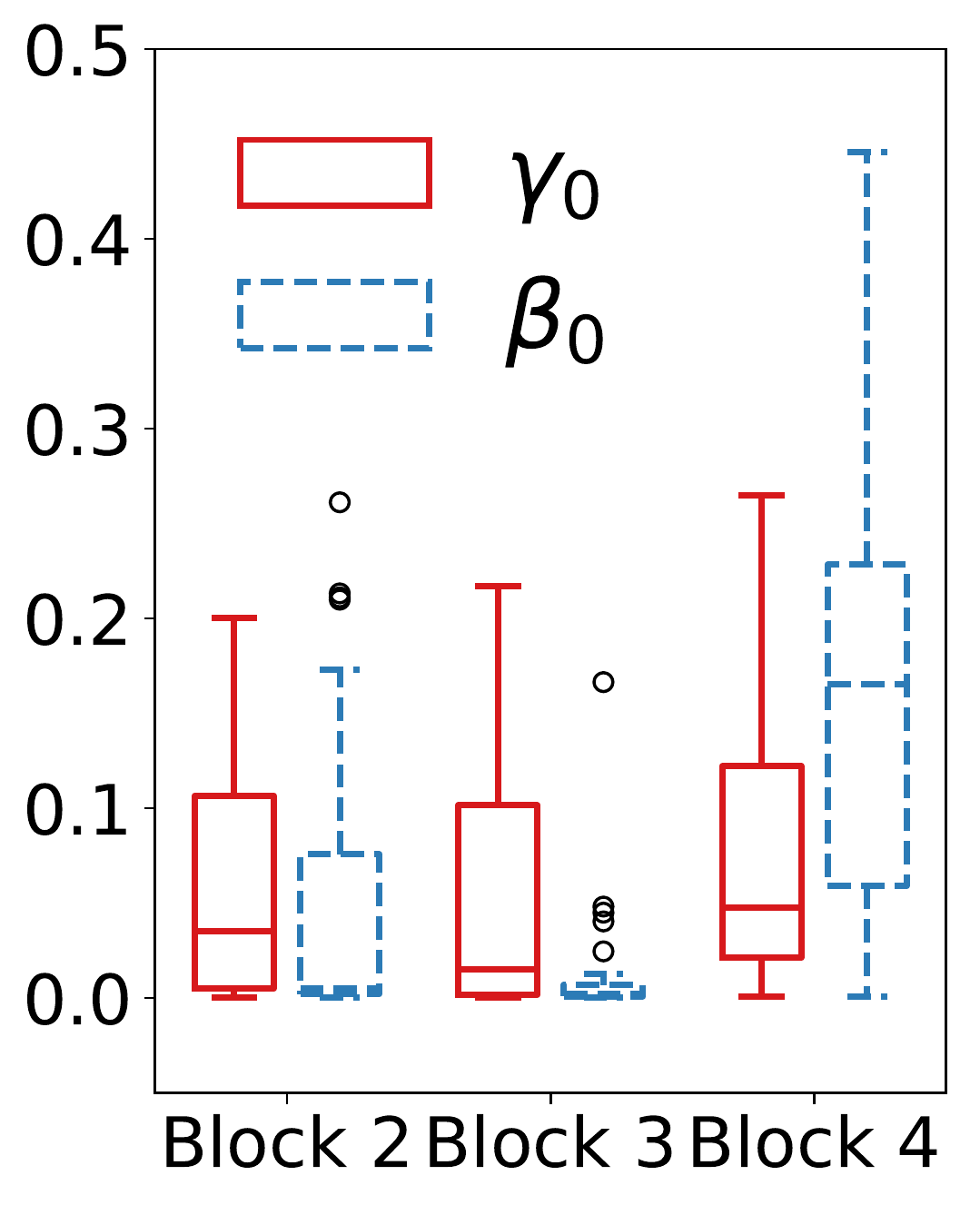}
}\hfill
  \subfloat[Analysis of weight matrix W\label{fig:generator}
  ]{\includegraphics[width=0.35\textwidth]{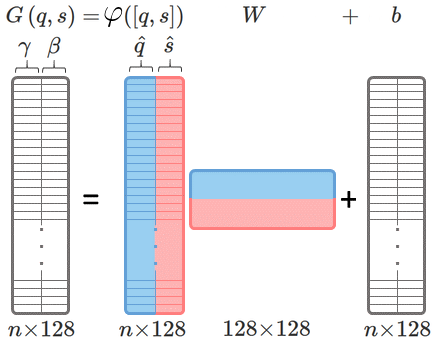}}\hfill
  \subfloat[Averaged norms per block\label{fig:block_norm_analysis}]{\includegraphics[width=0.3\textwidth]{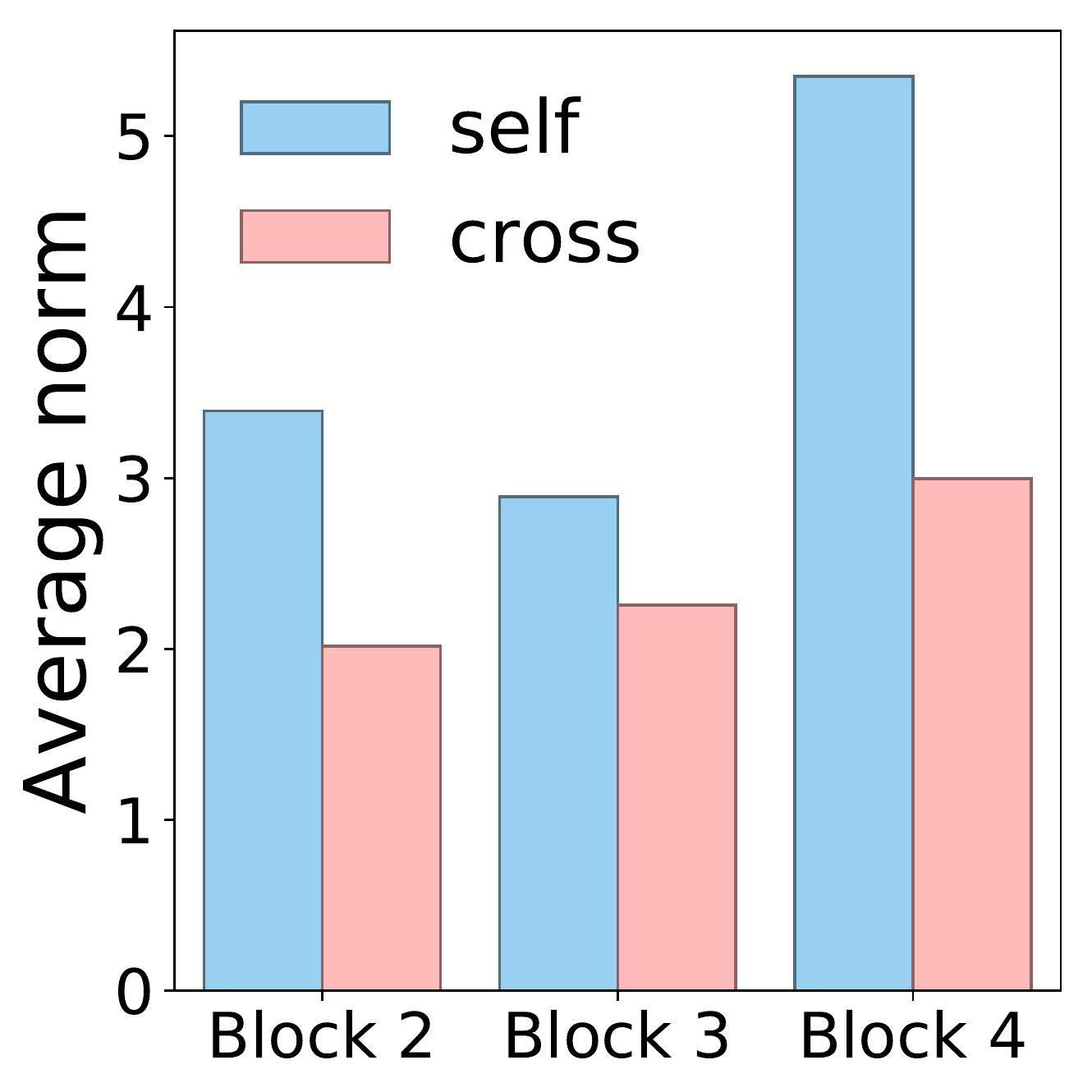}}
  \caption{Block-level modulation analysis: (a) distribution of absolute values of $\gamma_0$ and
    $\beta_0$ post-multipliers, (b) decomposition in self-modulation and cross modulation termsanalysis,  and (c) Average norms per block.}
  \label{fig:boxplots}
\end{figure}

\paragraph{The model applies cross-modulation at different levels of abstraction}

We assess whether the model learns to take advantage of
cross-modulation at various levels of abstraction in the network in
accordance to our motivating hypothesis. We perform an ablation study
similar to the one above where we randomly distort the modulation
selectively in different blocks (Table~\ref{tab:noise}, rows 2-4). Our
results suggest that the model has learned to take advantage of
cross-modulation at multiple levels of abstraction, but relies more
heavily on modulations in the second and fourth blocks. These results
match the absolute value distribution observed for the $\gamma_0$ and
$\beta_0$ post-multipliers (\autoref{fig:boxplots}), which regulate
the intensity of the modulation on a per channel basis.  We see that
the $\gamma_0$ distribution is slightly higher for blocks~2~and~4.

Note that in this analysis we do not {\em train} the model with fewer modulated
blocks; in that case, it is possible that it could learn to adapt to a different
architectural configuration. More experiments are needed to quantify the effect
of the specific number and location of modulated blocks on overall performance.

\paragraph{The model cross-modulates {\em and} self-modulates}

Recent work such as \cite{hu17squeeze} outlines the benefit of
{\em self-modulation} --- where the feature extraction pipeline interacts
with itself --- in a large-scale classification setting. The noncommutative
nature of our proposed cross-modulation mechanism allows both
{\em self-modulation} and {\em cross-modulation}. To disambiguate
between the two, we analyze the weight matrices of
the FiLM generator blocks. As Figure~\ref{fig:generator} shows, these
weight matrices can be viewed as the concatenation of two submatrices
responsible for self-modulation and cross-modulation, respectively.
Figure~\ref{fig:block_norm_analysis} shows the average weight matrix
column norm for the self- and cross-modulation submatrices at each
block. We see that self-modulation has a greater influence on the
predicted FiLM parameters, but that the model also takes advantage
of cross-modulation.

\section{Conclusion}

We have proposed a new architectural feature called {\em cross-modulation}
which allows support and query examples to interact at multiple levels of
abstraction and which can be used in conjunction with metric learning
approaches to few-shot classification. Initial experiments with a baseline
Matching Networks architecture show promising results in the {\em 5-way, 1-shot}
setting, and our analysis of the trained network shows that the model equipped
with cross-modulation learns to use it in meaningful ways. Interesting avenues
for future work include more thorough empirical verification (e.g. on more
datasets and network architectures), developing analogous cross-modulation
architectures for other methods such as Prototypical Networks, and addressing
scaling issues associated with the use of pairwise interactions at multiple
levels of abstraction.

\newpage

\bibliographystyle{unsrt}
\bibliography{biblio}

\end{document}